\pdfoutput=1
\documentclass[11pt]{article}

\usepackage[]{acl}
\usepackage{graphicx}
\usepackage{enumitem}
\usepackage{hyperref}
\usepackage{times}
\usepackage{latexsym}
\usepackage{bm}
\usepackage{xpatch}
\usepackage{lineno,microtype}
\usepackage[T1]{fontenc}
\usepackage[utf8]{inputenc}
\usepackage{booktabs}
\usepackage{multirow}
\usepackage{subfig}
\usepackage{colortbl}
\usepackage[table]{xcolor}

\usepackage{microtype}

\usepackage{arydshln}
\usepackage{amsmath}

\newcommand{\bigO}{\mathcal{O}}
\definecolor{lightyellow}{RGB}{255, 255, 204}

\newlength{\bibitemsep}
\newlength{\bibparskip}

\newcommand{\email}[1]{\href{mailto:#1}{#1}}
\title{Challenging Quadratic Attention---A Holistic View on the Rise of Alternative Language Model Architectures}

\author{Alexander M. Fichtl, {\bf Jeremias Bohn}, {\bf Josefin Kelber}, {\bf Edoardo Mosca} \and {\bf Georg Groh} \\
         Social Computing Group \\Technical University of Munich\\Boltzmannstraße 3, 85748, Garching, Germany\\\email{\{alexander.fichtl, jeremias.bohn, josefin.kelber, edoardo.mosca, georg.groh\}@tum.de}}

\begin{document}
\maketitle


\begin{abstract}
Transformers have dominated sequence processing tasks for the past seven years---most notably language modeling. However, the inherent quadratic complexity of their attention mechanism remains a significant bottleneck as context length increases. We review and distill the recent efforts to overcome this bottleneck, including advances in (sub-quadratic) attention variants, recurrent neural networks, state space models, and hybrid architectures. We critically analyze approaches regarding compute and memory complexity, benchmark results, and fundamental limitations to assess whether the dominance of pure-attention transformers may soon be challenged, which we consider possible, particularly in domain-specific and edge-device applications.
\end{abstract}


\section{Introduction}
\label{section:introduction}
The transformer architecture is a foundational breakthrough in \emph{Natural Language Processing} (NLP) \citep{transformer_vaswani_2017}, forming the backbone of most \emph{Large Language Models} (LLMs) \citep{gpt3} and is a reliable architecture choice for predictable performance scaling laws \citep{kaplan2020scalinglawsneurallanguage, hoffmann_training_2022}. Its self-attention mechanism \citep{atteniton_bahdanau_2015} projects inputs into \emph{queries} ($Q$), \emph{keys} ($K$), and \emph{values} ($V$), enabling efficient pairwise token interactions:
\[\mathrm{Attention}(Q, K, V) = \mathrm{softmax}\left(\frac{QK^\top}{\sqrt{d_k}}\right)V\]
\begin{figure}[t]
 \centering
 \includegraphics[width=0.48\textwidth]{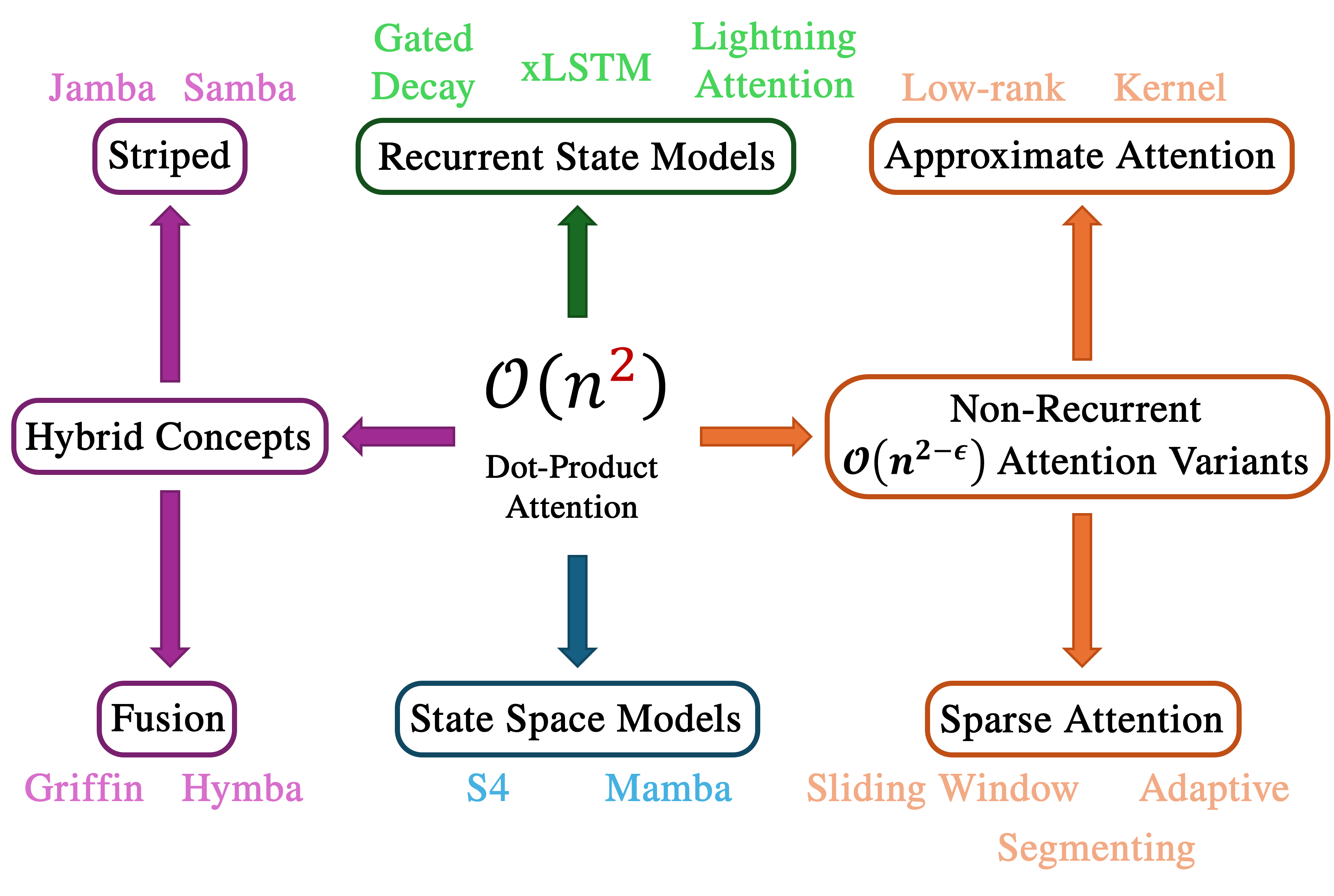}
 \caption{The four types of dot-product attention alternatives identified in our survey, including examples. We further divide hybrid concepts (striped and fusion hybrids), and sub-quadratic attention variants (approximate and sparse attention) each in two major classes.} \label{fig:abstract} 
\end{figure}
Despite providing direct $\mathcal{O}(1)$ paths between any pair of tokens, computing the full $n \times n$ attention matrix incurs $\mathcal{O}(n^2)$ time complexity, increasing latency and compute costs as the input length $n$ grows \citep{transformer_vaswani_2017}. While efficiency improvements to standard attention mostly focused on caching and memory layout (see Appendix \ref{subsection:moderno2}), its core problems have motivated research efforts into sub-quadratic sequence-modeling operators to replace attention, aiming to improve efficiency while retaining strong task performance. These include sub-quadratic attention variants \citep{linear_attention_katharopoulos_20}, \emph{Recurrent Neural Networks} (RNNs) \citep{beck_xlstm_2024}, \emph{State Space Models} (SSMs) \citep{mamba_gu_23, s4_gu_22}, and hybrids thereof \citep{de_griffin_2024}. 

This paper reports on the overarching narrative of developing alternatives to transformers, reviews the most impactful milestones and new primitives, and examines whether the transformer dominance may soon be challenged. Our main contributions are:
\begin{itemize}
\item[\textbf{(1)}] A review of the most relevant (sub)-quadratic attention variants, recurrent state models, SSMs, and hybrid architectures. An overview can be found in Figure~\ref{fig:abstract}.
\item[\textbf{(2)}] A comparative analysis of time and memory complexity for training and inference of sequence-modeling mechanisms, as well as reported benchmark results for SOTA models.
\item[\textbf{(3)}] A critical analysis of strengths, tradeoffs, and limitations, with an informed perspective on when and where pure attention-based transformers may be surpassed.
\end{itemize}
Our methodology is described in Appendix~\ref{sec:methodology}.

\section{Related Review Work}
While several recent and concurrent works overlap with aspects of our scope, they differ in focus and conclusions. \citet{schneider_what_2024} discusses hypothetical post-transformer architectures without restricting to sub-quadratic complexity or state-of-the-art performance. \citet{wang_beyond_2024} review approaches for handling longer input sequences, and \citet{tiezzi_back_2025} examine alternative architectures from a recurrent-processing perspective.

Several surveys provide overviews of efficient transformers and LLMs in general \citep{tay_efficient_2022, miao_towards_2023, 
huang_advancing_2023, wan_efficient_2024, miao_x-former_2024, tang_survey_2024}, but often emphasize linear attention variants when considering alternative architectures. Focused surveys address specific subgroups, such as SSMs \citep{somvanshi_s4_2025, ssmsurvey_wang_24} and recurrent models \citep{tiezzi_resurgence_2024}, or specific domains like computer vision \citep{patro_mamba360_2024} and time series forecasting \citep{kim_survey_2025}, whereas our emphasis is on NLP tasks and all sub-quadratic alternatives to attention-based models. Existing surveys also frequently give extensive full historical lineages of the discussed models \citep[e.g.,][]{sun2025speed}, while we focus on practical relevance in recent applications and research.
Finally, \citet{strobl_surveyexpressivity_2024} provide an overview of works on transformer expressivity, which relates to our discussion of architectural limitations in Section \ref{section:Limitations}.

\section{Non-Recurrent \texorpdfstring{\boldmath$\mathcal{O}(n^{2-\epsilon})$}{Sub-Quadratic} Attention} \label{subsection:LinearAttention} 
Categorizing sub-quadratic attention alternatives is challenging due to overlapping ideas and mechanisms. We organize them as non-recurrent attention variants, recurrent state models, SSMs, and Hybrids according to their main design motivation, though some 
fall into several categories.
\subsection{Approximate Attention} 
\label{paragraph:ApproximateAttention} 
These mechanisms, including linear attention, reduce computational cost by using approximations such as kernel functions or low-rank factorization. Kernel-based linear attention reformulates self-attention as a linear dot-product in feature space, achieving $\mathcal{O}(n)$ complexity \citep{linear_attention_katharopoulos_20, shen_efficient_2024}. However, a poorly chosen kernel can result in reduced expressivity. Sequential cumulative summation can also slow inference in causal settings, as seen in the Performer \citep{choromanski_rethinking_2020}. Low-rank methods---e.g., Linformer \citep{wang_linformer_2020}---similarly achieve $\mathcal{O}(n)$ complexity, but their effectiveness depends on the selected rank.
Notably, the Performer performs worse on autoregressive generation than for masked language modeling (MLM), and the Linformer is not applicable to decoder-based modeling in general. 
The \emph{Attention Free Transformer} (AFT) \citep{zhai_attention_2021} replaces dot product self-attention by learned position biases. The biases are added to the keys and values, and the result is multiplied by the query element-wise, which is an advantage for large model sizes. Later, Hyena \citep{poli_hyena_2023} generalizes the work of AFT by combining multiplicative element-wise gating with implicit long convolutions. \footnote{Hyena, however, includes a recurrent operator with a time complexity of $\mathcal{O}(NL\log_2 L)$ \citep{poli_hyena_2023}}

However, while these variants are important foundational works, they are generally no longer competitive with more recent architectures.
Recent competitive variants, such as ReGLA \citep{lu_regla_2025}, Hedgehog \citep{zhang_hedgehog_2024}, and RoFly \citep{ro_fly_2025}, improve efficiency while also enhancing expressivity. Log-linear attention \citep{guo2025loglinearattention} extends linear attention through a logarithmically growing set of hidden states, providing a trade-off between efficiency and expressiveness. 

\subsection{Sparse Attention} 
\label{paragraph:SparseAttention} 
Sparse attention mechanisms focus computation on a subset of the sequence using fixed or learnable patterns. Sparse Transformers \citep{child_generating_2019} pioneered sparse factorizations of the attention matrix, reducing complexity to $\mathcal{O}(n\sqrt{n})$\@. Local (sliding window) attention restricts computation to a window around each token and is often paired with global attention, as in Longformer \citep{beltagy_longformer_2020}, to regain expressivity by allowing selected tokens to attend globally. Other variants, such as strided or random patterns, are often combined \citep[e.g.,][]{zaheer_big_2020}. While some sparse patterns achieve $\mathcal{O}(n)$ time and memory complexity, they may underperform on tasks requiring fine-grained global dependencies and often require task-specific tuning. Learnable and adaptive sparsity patterns \citep[e.g.,][]{correia_adaptively_2019} can address these limitations. 
More recent examples of sparse attention are MoBA \citep{lu_moba_2025} or NSA \citep{yuan_NSA_2025}: NSA uses hardware-aligned sparse attention kernels and a learned gating mechanism to combine sliding attention, compressed attention for coarse-grained patterns, and selected attention for important token blocks \citep{yuan_NSA_2025}. MoBA is inspired by \emph{Mixture-of-Experts} (MoE) systems, divides the context into blocks, and uses a dynamic gating mechanism to selectively route query tokens to the KV blocks most relevant to them \citep{lu_moba_2025}.

\section{Recurrent State Models}
\label{subsection:RecurrentModels}
\emph{Recurrent Neural Networks} (RNNs) process sequences by maintaining a fixed-size state updated at each time step, allowing them to model temporal dependencies \citep{rnns_yu_19}\@. \emph{Long Short-Term Memory} (LSTM) networks \citep{lstm_hochreiter_97} mitigate the vanishing gradient problem through a complex gating mechanism, while \emph{Gated Recurrent Units} (GRU) \citep{gru_cho_14} offer a simpler alternative with similar performance and lower computational cost.
RNN variants offer linear autoregressive generation, but suffer from (1) varying degrees of vanishing/exploding gradients, (2) limited training parallelism, and (3) lack of expressivity due to a representation state not scaling with context length \citep{rnns_yu_19}.

While the following models are not RNNs per se, they recurrently update a state in the form of a matrix (in contrast to single vectors in classic RNNs) where the influence of earlier inputs decays over time, thus a formulation as such a model is possible. These successors partly mitigated the limitations named above, starting with \citet{linear_attention_katharopoulos_20}, which introduce linear attention.

\subsection{Gated Attention Models}
Instead of attending to a full key-value matrix of previous outputs, gates can recombine a fixed context state that tokens attend to with new outputs. The Retentive Network (RetNet) \citep{sun2023retentive} builds on linear attention, improving the performance by applying exponential decay to the hidden state before the RNN update.
It introduces the retention mechanism for sequence modeling, whose parallel representation (for efficient training) resembles self-attention but replaces the softmax operation by a Hadamard product, data-independent exponential decay, and GroupNorm. This enables a recurrent representation and consequently low-cost $\mathcal{O}(1)$ inference per token.
Moreover, a chunkwise recurrent representation combines parallel encoding within chunks with recurrent summarization to efficiently model long sequences with linear complexity. 


Successively, Gated Linear Attention (GLA) \citet{yang_gatedLinearAttention_2024} introduces data-dependent gates and the hardware-efficient algorithm FlashLinearAttention. This enabled GLA Transformers to perform competitively with full attention for small-scale language models at the time and came with no significant perplexity degradations when using a model trained on 2000 tokens on sequences longer than 20000 tokens, which demonstrates their effectiveness at length generalization.
Since the gating computation per token depends only on the current token and the learnable parameters, it remains parallelizable.
 
DeltaNet \citep{pmlr-v139-schlag21a, yang_parallelizingLinear_2025} is a linear transformer variant that retrieves and updates a value vector associated with each key using an update rule similar to the delta rule. DeltaNet employs a \emph{diagonal plus low-rank} (DPLR) state-update mechanism similar to S4, enabling efficient parallelization across the temporal dimension and significantly improving training efficiency. Gated DeltaNet \citep{yang2025gated} builds upon this and introduces the gated delta rule, which integrates gating for adaptive memory control. Similarly, Goose (RWKV-7) \citep{peng_rwkv-7_2025}, the latest RWKV version, integrates a generalized delta rule, vector-valued gating, in-context learning rates, and a relaxed value replacement rule. The initial RWKV-4 \citep{peng_rwkv_2023} uses token shift and builds on AFT \citep{zhai_attention_2021} by using channel-wise time decay vectors instead of global interaction weights. See \citet{li_survey_2025} for a detailed overview.


\subsection{Lightning Attention (LA2)} 
Lightning Attention-2 \citep{qin_lightning_2024} divides attention into intra-block (standard attention) and inter-block (linear attention via kernel tricks) computations. This divide-and-conquer strategy addresses the slow training of causal linear attention---caused by sequential cumulative sums---by combining efficient intra-block processing with fast, kernel-based inter-block calculations. LA2 keeps a fixed-size hidden state and is considered a data-independent decay variant.
While the forward pass of LA2 resembles RetNet's chunk-wise retention algorithm \citep{sun2023retentive}, LA2 additionally includes the backward pass, incorporates IO-aware optimizations from FlashAttention, and enhances GPU performance through tiling. Both forward and backward passes have time complexity $\mathcal{O}(nd^2)$ \citep{qin_various_2024}\@. 
Although TransNormerLLM \citep{qin_transnormerllm_2023} is tailored for Lightning Attention, LA2 itself is model-agnostic. MiniMax-01 \citep{li2025minimax} uses LA2 and reports that, under a fixed compute budget, it enables more parameters and tokens, achieving lower loss than standard softmax attention.

Other models that belong in this section, namely the xLSTM \citep{beck_xlstm_2024} and HGRN families \citep{qin_hierarchically_2023}, are described in Appendix~\ref{sec:honorable_mentions}.

\section{State-Space-based Models}
\label{subsection:SSSMs}
\emph{State Space Models} (SSMs), originally from control theory for modeling dynamic systems via state variables \citep{kalman1960new}, have emerged as promising sub-quadratic alternatives to transformers. A key aspect is their dual perspective: a recurrent formulation enables $\mathcal{O}(n)$ inference, while a convolutional view allows for $\mathcal{O}(n\log(n))$ training via efficient FFT-based convolutions. 
Note that the models are listed in this separate subsection, not as a concept distinct from recurrent state models, but for their importance in current research.
\subsection{Structured SSMs} Structured SSMs impose a specific mathematical structure---such as low-rank or diagonal-plus-low-rank forms---on state transition and input matrices, enabling efficient and expressive modeling of long-range dependencies. S4 \citep{s4_gu_22} introduces the use of a \emph{Highly Predictive Polynomial Projection Operator} (HiPPO) matrix for initializing the state transition. This approach enables the construction of global convolution kernels that can efficiently encode long-term dependencies. At the time of release, S4 matched the performance of transformers \citep{s4_gu_22}\@. S5 \citep{s5_smith_23} simplifies and extends S4 by replacing its diagonal block structure with dense matrices. 
Additionally, S5 leverages an efficient parallel scan, removing the need for S4’s convolutional and frequency domain computations and streamlining kernel computation.
    
\subsection{Selective SSMs}
Mamba \citep{mamba_gu_23} advances SSMs by replacing fixed transition matrices with input-dependent functions, increasing flexibility and expressivity. Its core is the Mamba block, which combines the ideas of H3 \citep{h3_hungry} and gated MLP blocks by adding a convolution and an SSM to the main branch of the gated MLP\@. Efficient implementation is achieved via kernel fusion, parallel scan, and recomputation.

Mamba2 \citep{mamba2_dao_24} further unifies structured SSMs with attention mechanisms, enabling the application of transformer-style optimizations. It uses modified Mamba blocks for tensor parallelism and introduces the \emph{State Space Dual} (SSD) layer as the inner SSM, which, in its recurrent form, is a selective SSM with single-input single-output structure. This design slightly reduces expressivity but significantly improves training efficiency on modern accelerators.
\section{Hybrids}

\begin{figure}[ht]
    \centering
    \subfloat[Striped Hybrid]{
    \includegraphics[height=.31\textheight]{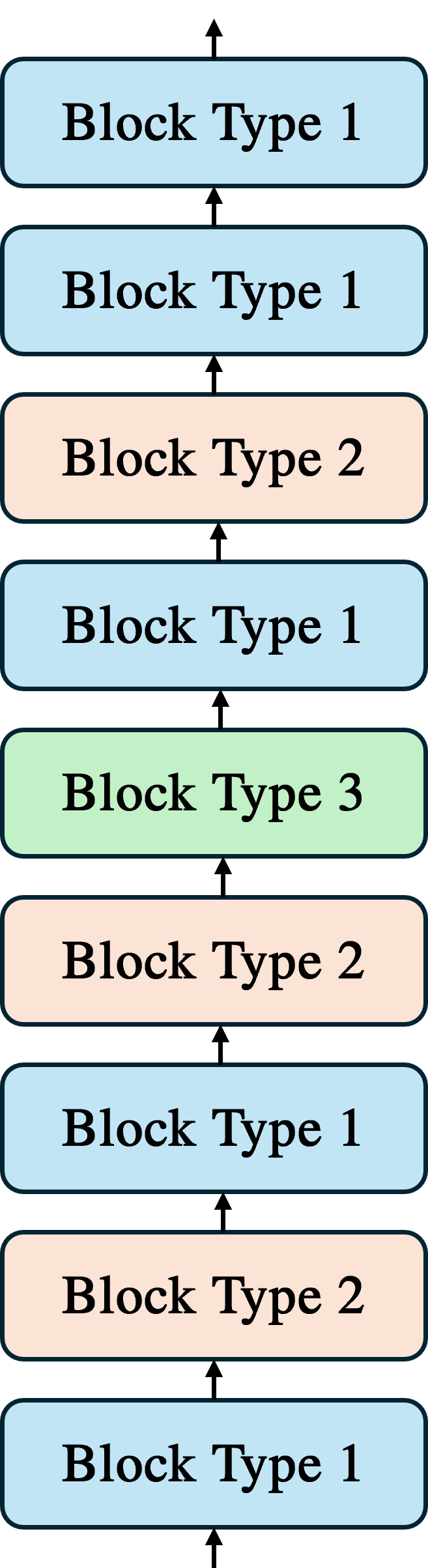}
    }
    \quad
    \subfloat[Fusion Hybrid]{
    \includegraphics[height=.31\textheight]{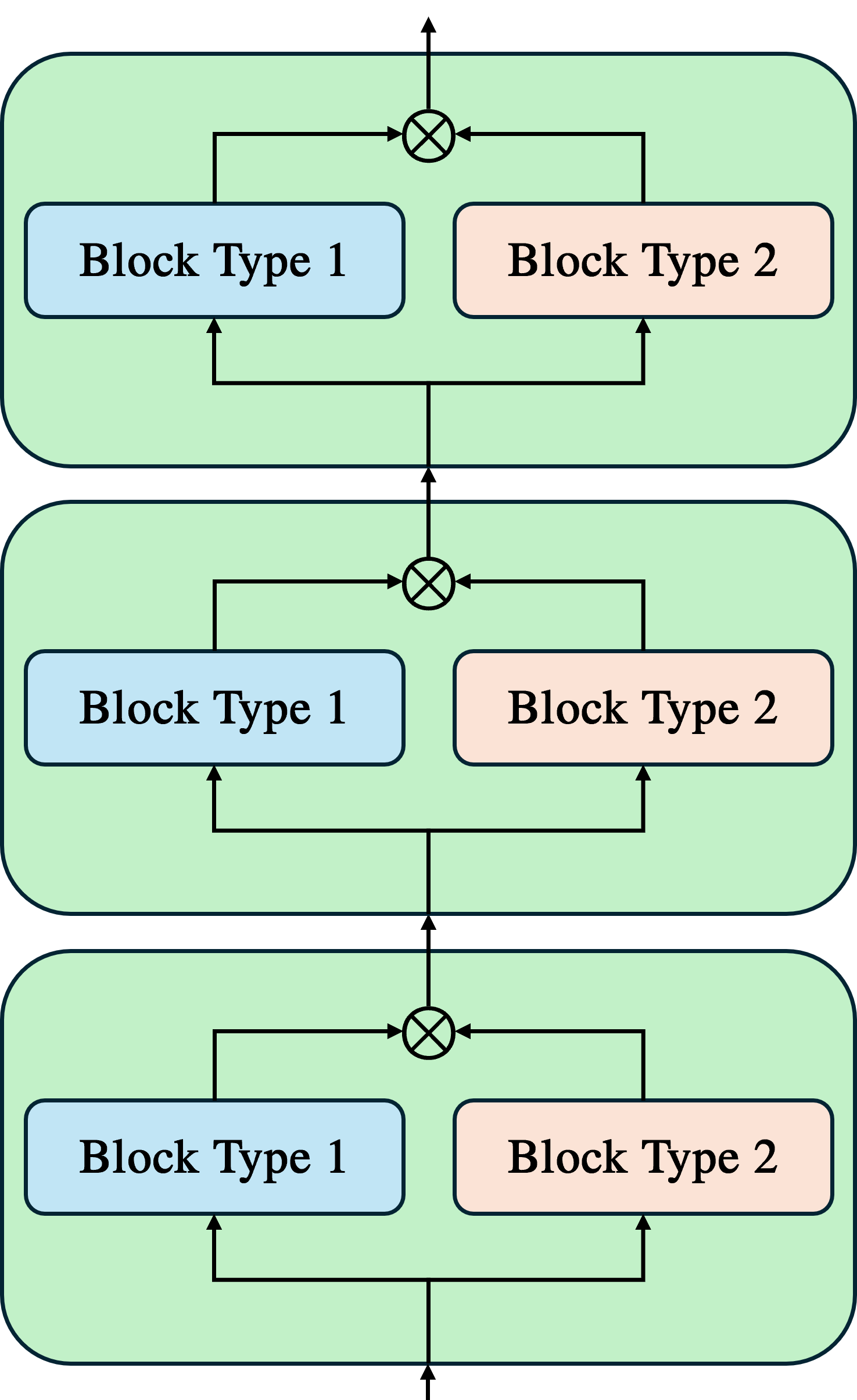}
    }
    \caption{Different types of hybrids. \textbf{(a)}: block types using different primitives are connected in series. \textbf{(b)}: block types are connected in parallel.}
    \label{fig:hybrids}
\end{figure}

Hybrid architectures combine different primitives---such as SSMs, attention, and RNNs---to leverage their strengths while mitigating the limitations of individual approaches (see Sections~\ref{subsection:LimitationsofAttention} and \ref{subsection:limits_of_subq_alternatives}). Such hybrids are usually of a striped (i.e., alternating primitives in series) or a fusion nature (i.e., primitives are calculated in parallel, combining their outputs). See Figure ~\ref{fig:hybrids} for reference.

\subsection{\texorpdfstring{\boldmath$\mathcal{O}(n^2)$}{Quadratic} Hybrids}
\label{subsection:quadraticHybrids}
\paragraph{SSM + Quadratic Attention}
Several works demonstrate that combining SSM and attention layers often outperforms using either one alone: For instance, \citet{mamba2_dao_24} show that integrating SSD layers, attention, and MLPs can surpass pure Transformers or Mamba-2\@. Jamba \citep{lieber_jamba_2024} merges transformer, Mamba, and MoE layers into a striped hybrid, achieving performance comparable to Llama-2 70B or Mixtral, but has 2–7x longer context windows, 3x higher throughput, fewer total parameters (52B, 12B active), and reduced KV cache memory (32GB vs. 4GB for 256K tokens). Other examples are MambaFormer~\citep{park_mambaformer_2024}, another striped hybrid, and Hymba~\citep{dong_hymba_2025}, combining fusion and striped hybrid patterns.

\paragraph{Lightning Attention + Quadratic Attention}

\citet{li2025minimax}'s MiniMax-01 series combines lightning attention with an MoE approach. To address LA2’s limited retrieval, their Hybrid-lightning architecture replaces LA2 with $\mathcal{O}(n^2)$ attention every eight layers, resulting in a striped hybrid. MiniMax-Text-01 was competitive with SOTA models like GPT-4o or Claude-3.5-Sonnet at the time of release, supporting context windows up to 1M tokens during training and 4M during inference at reasonable cost. However, it still struggles to follow multilevel instructions due to sparse training data.

\paragraph{Gated Attention + Quadratic Attention}

Kimi Linear \citep{kimi_linear} uses Kimi Delta Attention (KDA), a linear attention mechanism that refines the gated delta rule with fine-grained gating. The proposed architecture has three 3 KDA layers for every global attention (MLA) layer.

\subsection{\texorpdfstring{\boldmath$\mathcal{O}(n^{2-\epsilon})$}{Sub-Quadratic} Hybrids}
\label{subsection:SubquadraticHybrids}
\citet{de_griffin_2024} propose the \emph{Real-Gated Linear Recurrent Unit} (RG-LRU), a gated LRU \citep{orvieto_resurrecting_2023} variant without complex transformations in the recurrence, as these do not improve language modeling in practice. RG-LRU, a fusion hybrid of local attention and linear recurrence, is used for sequence mixing in a recurrent block, replacing MQA.
Griffin, using RG-LRU, achieves higher inference throughput and lower latency on long sequences than MQA Transformers \citep{de_griffin_2024}\@. On benchmarks, Griffin-3B outperforms Mamba-3B, and Griffin-7B and 14B are competitive with Llama-2 despite using much less training data. Griffin is also used as the base for RecurrentGemma \citep{botev_recurrentgemma_2024}.

Another notable sub-quadratic hybrid is Samba \citep{ren_samba_2025}, a striped hybrid combining sliding window attention and Mamba/SSM layers.
\begin{table*}[ht]
    \centering
    \resizebox{\textwidth}{!}{
        \begin{tabular}{lcccccc}
            \toprule
            \multicolumn{1}{c}{\multirow{2}{*}{\textbf{Method}}} 
            & \multicolumn{3}{c}{\textbf{Training}} 
            & \multicolumn{2}{c}{\textbf{Inference}} \\
            \cmidrule(lr){2-4}\cmidrule(lr){5-7}
            & \textbf{Time} & \textbf{Space} & \textbf{Parallel} 
            & \textbf{Time} & \textbf{Space} \\
            \midrule
            FFT-Convolution         & $\bigO(Bnd\log(dn))$ & $\bigO(Bnd)$ & Yes & $\bigO(nd\log(nd))$ & $\bigO(nd)$\\
            RNN                 & $\bigO(Bnd^2)$ & $\bigO(Bnd)$ & No & $\bigO(d^2)$\footnotemark[2] & $\bigO(nd)$ \\
            Vanilla Transformer & $\bigO(B(n^2d + nd^2))$ & $\bigO(B(n^2 + nd))$ & Yes & $\bigO(n^2d + d^2n)$ & $\bigO(n^2+nd)$ \\
            LSH (Reformer)    & $\mathcal{O}(Bd^2n \log n)$ & $\mathcal{O}(Bn \log n + Bnd)$ & Yes & $\mathcal{O}(d^2n \log n)$ & $\mathcal{O}(n \log n + nd )$ \\
            FAVOR+ (Performer)   & $\mathcal{O}(Bnd^2 \log d)$ & $\mathcal{O}(Bnd \log d + Bd^2 \log d)$ & Yes & $\mathcal{O}(nd^2 \log d)$ & $\mathcal{O}(nd \log d + d^2 \log d)$ \\
            Linear Transformer  & $\bigO(Bnd^2)$ & $\bigO(B(nd+d^2))$ & Yes & $\mathcal{O}(nd^2)$ & $\mathcal{O}(nd + d^2)$ \\
            Lightning Attention & $\bigO(Bnd^2)$ & $\bigO(B(nd+ d^2))$ & Yes & $\bigO(nd^2)$ & $\bigO(nd+ d^2)$ \\
            Channel-wise              & \multirow{2}{*}{$\bigO(Bnd^2)$} & \multirow{2}{*}{$\bigO(Bnd)$} & \multirow{2}{*}{Yes} & \multirow{2}{*}{$\mathcal{O}(nd)$} & \multirow{2}{*}{$\mathcal{O}(d)$} \\
            AFT (RWKV-4)&&&&&\\
            Hyena-3             & $\bigO(Bnd\log(dn))$ & $\bigO(Bnd)$ & Yes & $\bigO(nd\log(n+d))$ & $\bigO(nd)$ \\
            S4                  & $\bigO(Bnd\log(dn))$ & $\bigO(Bnd)$ & Yes & $\bigO(d^2)$ & $\bigO(nd)$\\
            Mamba\footnotemark[3] & $\bigO(B(nd^2 + nd\log(nd)))$ & $\bigO(Bnd)$ & Yes & $\bigO(nd^2 + nd\log(nd))$ & $\bigO(nd)$ \\
            \bottomrule
        \end{tabular}
    }
    \caption{Overview on time \& space complexities for training on a single batch and inference of a single token of different sequence-modeling mechanisms. $n$: sequence length; $d$: hidden dimension; $B$: batch size}\label{tab:complexities}
\end{table*}
\subsection{Novel Architecture Design Concepts}
\paragraph{Memory System Design}
Recent models increasingly integrate several memory types \citep{irie_blending_2025, nunez_span_2025}\@. Titans \citep{behrouz_titans_2024} introduce meta in-context neural long-term memory, storing surprising data at test time, and combine core attention-based short-term, neural long-term, and persistent task memory modules.

B'MOJO \citep{zancato_bmojo_2025} generalizes transformers and SSMs by blending permanent, short-term, fading, and long-term memories, with a sliding attention mechanism to aggregate information. Both models show good results versus transformers on several benchmarks (see Table~\ref{tab:benchmarks}).
\paragraph{Tailored Architecture Search}
\citet{thomas2024starsynthesistailoredarchitectures}'s STAR framework unifies popular sequence model architectures under the theory of \emph{Linear Input-Varying systems} (LIVs), creating a larger and more structured search space for model design. Given target metrics such as cache size, perplexity, or device latency, STAR uses gradient-free evolutionary algorithms to automatically search the LIV space and generate architectures optimized for several objectives, outperforming highly-tuned transformer and hybrid models on various quality and efficiency frontiers. A recent model realized through STAR (with slight modifications) is the edge model LFM2 \citep{lfm2}.

Another example of architectural search is PostNAS \citep{gu2025jet}, which enables an exploration of attention block designs building on pre-trained transformer models.

\section{Complexity and Benchmark Analysis}
\label{subsection:QuantitativeAnalysis}
Moving beyond the qualitative analysis in the previous sections, this section focuses on quantitative results and directly compares model architectures in terms of complexity and benchmark performance.
\subsection{Complexity Comparison} 
We compare the complexities of selected sequence-modeling mechanisms in Table~\ref{tab:complexities}\@. It is important to note that these complexities are sometimes dominated by feed-forward neural networks in the full model, e.g., in S4, which have a time complexity of $\bigO(nd^2)$\@. Except for RWKV, which can process a single query at a time at inference, models are lower bounded on memory complexity by storing the sequence in its entirety. Many of these algorithms rely on projections, thus requiring at least $\bigO(nd^2)$ operations, often serving as an upper bound for time complexity.
Another major influence on time complexity is the use of FFT convolutions, as used in SSM-based models for training, which requires $\bigO(nd\log(dn))$ computational steps, binding the algorithm to log-linear time.
\footnotetext[2]{Assuming the sequence has been processed already, only necessary once}
\footnotetext[3]{We consider an entire Mamba layer here, including projections}
\setcounter{footnote}{3}
\subsection{Benchmark Performance}
\label{subsection:benchmark_performance}
In Table~\ref{tab:benchmarks}, we provide a performance comparison of previously mentioned sub-quadratic models with recent high-performing quadratic attention models. We chose a frequently used configuration variety: two table sections comparing models with parameter sizes of 0.7-1.5B and 14-70B (total parameter count for MoE models) across eight prominent benchmarks. For the model and benchmark sources, see Appendix~\ref{sec:benchmarking_details}.
In the 0.7-1.5B range, several edge models compete for the top scores. In particular, Samba and RWKV7-World3 significantly outperform the full attention Llama 3.2 and Qwen2.5 in several instances. In the midrange (14-70B), no pure sub-quadratic models are present anymore; merely the hybrids Griffin and Jamba remain, with only the latter realistically competing with Qwen2.5 and Llama3.1. In the evaluation of frontier (100B+) models, we refer to the LMsys Chatbot Arena \citep{lmsys_chatbot} instead of a custom-made table. Across all benchmarks (accessed on 2025-10-02), only MiniMax-Text-01 \citep{li2025minimax} appears in the top-30 ranking once\footnote{There have been new developments between initial and camera-ready submission, some Hybrids actually reached the top 20 of the \textit{Text}, and top 5 of the \textit{WebDev} leaderboard, see Section~\ref{subsection:Outlook}.}, specifically in the \textit{WebDev} Arena leaderboard. In the top 10, no model is known to be built on alternative architectures.

Note that the benchmark comparison should only be taken as a rough overview. Public leaderboards and benchmarks are highly volatile, and scores reflect only the status at the given timestamp. Moreover, unstandardized benchmarking makes comparing architectures throughout the literature difficult (see Appendix~\ref{sec:benchmarking_details} and Limitations for details).

\begin{table*}[ht]
\footnotesize
\centering
\begin{tabular}{lcccccccc}
\toprule
\textbf{Model} & \multicolumn{7}{c}{\textbf{Benchmark Selection}} \\
\midrule
\textit{0.7-1.5B} & Size & \textbf{MMLU} & \textbf{LMB} & \textbf{ARC-E} & \textbf{ARC-C} & \textbf{Wino.} & \textbf{Hella.} & \textbf{PIQA} \\
\midrule
Titans-MAG & 760M & - & 41.0 & 68.2 & 36.2 & 52.9 & 48.9 & 70.3 \\
\textbf{Griffin} & 1B & 29.5 & - & 67.0 & 36.9 & 65.2 & 67.2 & \textbf{77.4} \\
Llama3.2* & 1B & 32.1 & 63.0 & - & - & 60.7 &  63.7 & -\\
GLA & 1.3B & - & 46.9 & 57.2 & 26.6 & 53.9 & 49.8 & 71.8 \\
HGRN2 & 1.3B & - & 49.4 & 58.1 & 28.1 & 52.3 & 51.8 & 71.4 \\
\underline{Mamba2} & 1.3B & - & \underline{65.7} & 61.0  & 33.3 & 60.9 & 59.9 & 73.2 \\
xLSTM[1:0] & 1.3B & - & 57.8 & 64.3 & 32.6 & 60.6 & 60.9 & 74.6 \\
BMoJo-Fading & 1.4B & - & 45.4 & 52.3 & 26.6 & 53.3 & 46.0 & 70.0 \\
\textbf{RWKV7-World3} & 1.5B & 43.3 & \textbf{69.5} & \underline{78.1} & 44.5 & \underline{68.2} & \textbf{70.8} & \underline{77.1} \\
\textbf{Qwen2.5}* & 1.5B & \textbf{60.9} & 63.0 & 75.5 & \textbf{54.7} & 65.0 & \underline{67.9} & 75.8 \\
\textbf{Samba} & 1.7B & \underline{48.0} & - & \textbf{79.3} & \underline{48.2} & \textbf{72.9} & 49.7 & \underline{77.1} \\

\midrule
\textit{14-70B} & Size & \textbf{MMLU} & \textbf{BBH} & \textbf{GSM8K} & \textbf{ARC-C} & \textbf{Wino.} & \textbf{Hella.} & \textbf{HumanEval} \\
\midrule
Griffin & 14B & 49.5 & - & - & 50.8 & 74.1 & 81.4 & -  \\
\underline{Qwen}* & 14B & \underline{79.7} & 78.2 & 90.2 & 67.3 & 81.0 & 84.3 & \underline{56.7} \\
Jamba & 52B & 67.40 & 45.40 & 59.9 & 64.40 & 82.5 & 87.1 & 29.30  \\
Mixtral* & 56B & 70.6 & - & 60.4 & 59.7 & 77.2 & 84.4 & 40.2  \\
\textbf{Llama3.1}* & 70B & 79.5 & \underline{81.0} & \underline{95.1} & \underline{68.8} &  \textbf{85.3} & \textbf{88.0} & 48.2  \\
\textbf{Qwen2.5}* & 72B & \textbf{86.1} & \textbf{86.3} & \textbf{95.8} & \textbf{72.4} & \underline{83.9} & \underline{87.6} & \textbf{59.1}  \\
\bottomrule
\end{tabular}
\caption{Performance comparison of recent pure quadratic attention LMs (highlighted with *) and subquadratic models of similar size. Best results for each parameter category are marked in \textbf{bold}, second-best results are \underline{underlined}\@. Model names are in bold or underlined when they scored first or second at least once. Results are accuracy-based and rounded to one decimal point. For sources, see Section~\ref{sec:benchmarking_details}}
\label{tab:benchmarks}
\end{table*}

\section{Fundamental Architectural Limitations}
\label{section:Limitations}
Both quadratic attention and sub-quadratic architectures face fundamental limitations that cannot be overcome by scaling parameters or training. In this section, we discuss these inherent restrictions. General limitations of language models \citep[e.g.,][]{Wheeler_proceduralmemory_2025} are beyond this survey's scope.
\subsection{Limitations of Attention}
\label{subsection:LimitationsofAttention}
\paragraph{General Theoretical Expressivity}
The standard transformer forward pass belongs to the log-time uniform $TC^0$ circuit complexity class \citep{log-Precision-Transformers}\@. This fundamentally limits its ability to simulate finite automata or solve graph connectivity---necessary for state tracking and multi-step reasoning \citep{merrill_depth_2025}\@. In practice, such tasks are tractable for short contexts (e.g., by using transformers of depth $\bigO(\log C)$ for context length $C$), but remain infeasible for unbounded inputs under standard complexity assumptions.  To scale up these capabilities, the model dimension must grow with the task complexity, as is also highlighted in related work \citep{hahn-2020-theoretical, sanford_representational_2023}.

Allowing intermediate steps, i.e., \emph{Chain of Thought} (CoT) \citep{wei_cot_2022}, increases transformer expressivity w.r.t.\ the number of steps. \citet{li_chain_2024} show that with $T$ CoT steps, constant-depth transformers with $\bigO(\log n)$ embeddings can solve any problem solvable by boolean circuits of size $T$\@. Additionally, \citet{qiu_ask_2025} prove that prompting is Turing-complete: for any computable function, a finite-size transformer can compute it with an appropriate prompt. However, these enhancements also introduce new drawbacks, as shown by \citet{amiri_lowerboundscot_2025, peng_limitationstransformer_2024, saparov_search_2025}.

\paragraph{Length Generalization}
Transformers struggle to extrapolate, i.e., to generalize from shorter training context sizes to longer test sequences. In addition to being limited by memory constraints, the transformer architecture has fundamental length-generalization limits caused by positional encodings \citep{kazemnejad2023the}\@. While transformers without position encodings (NoPE) seem to be an alternative and work for longer sequences than explicit encodings, they still impose a context length limit \citep{wang-etal-2024-length}\@. 

Building upon \citet{huang2025_generalization}'s framework to analyze length generalization, \citet{veitsman2025borntransformertransformer} show that, if pretraining is done right, certain capabilities w.r.t.\ length generalization of transformers can be improved, but fundamental limitations persist. For models like SSMs and B’MOJO, the length generalization is instead limited by the capacity of the recurrent state.

For \citet{huang2025_generalization}' s framework or more details on limitations of attention, see Appendix~\ref{sec:additional_limitations}.

\subsection{Limitations of Sub-Quadratic Alternatives}
\label{subsection:limits_of_subq_alternatives}
Sub-quadratic architectures share some limitations with quadratic attention---e.g., SSMs are also in the complexity class $TC^0$\@ \citep{Merrill_illusion_2024}. Furthermore, these models introduce additional new challenges due to the inherent difficulty of compressing sequence context into a reduced state.

This finite state capacity has strong implications for ``lookup table'' tasks (e.g., MQAR \citep{MQAR}, $\text{hop}_k$ \citep{sanford_transformers_2024}), where such information is part of the input, as SSMs cannot recall an arbitrary amount of information previously seen \citep{arora_tradeoff_2024,de_griffin_2024, jelassi_ssmsCopying_2024}, even though recent work \citep{grazzi_mambaInContextLearning_2024} shows that some improvements can be made, as seen in Mamba \citep{mamba_gu_23}.

A similar problem occurs in linear RNNs, which are highly sensitive to the context order, making prompt engineering critical---selection and recall become harder as input order varies \citep{sutskever2014, arora_just_2024}\@. RNNs require $\Omega(N)$ space for reliable recall \citep{arora_tradeoff_2024}, and constant-memory models cannot perform associative recall
or solve tasks like $q$-sparse averaging or copying, unlike shallow transformers \citep{sanford_transformers_2024, jelassi_ssmsCopying_2024, wen_rnns_2025}.

\citet{backurs_editdistance_2018} prove that under SETH (which implies $P\neq NP$), edit distance cannot be computed in subquadratic time, setting a fundamental limit on sequence comparison efficiency for any such architecture. Under the same assumption, \citet{alman_fundamental_2025} show that document similarity tasks inherently require quadratic~time.
\subsection{Implications}
The limitations applying to alternative architectures mostly subsume the limitations of transformers. This implies that while sub-quadratic alternatives significantly enhance efficiency and lower computational costs, they do not fundamentally surpass transformers in theoretical expressivity.

\section{Discussion}
\label{section:Discussion}
In this section, we synthesize insights from our review to discuss whether sub-quadratic and hybrid alternatives start claiming meaningful territory.
\subsection{Current Landscape} 
Despite the reviewed advances in alternative architectures, at the time of writing, most frontier general-purpose models strongly rely on full attention mechanisms. No model scoring in the top 10 on LLMSys \citep{lmsys_chatbot} is known to be sub-quadratic or a hybrid, showing that the ``Transformer++'' remains the default choice when compute is not a limiting factor. We have also seen that full attention is free from many limitations that apply to alternative architectures (Section \ref{subsection:limits_of_subq_alternatives}), adding to the extent of their superiority. 

However, the picture changes for edge models, where compute, memory, and latency are tightly bound, and alternative architectures have gained substantial traction. Especially hybrids, such as Samba \citep{ren_samba_2025} or RWKV7 \citep{peng_rwkv-7_2025}, offer favorable inference properties. They can meet resource constraints by offloading local or intermediate computations to more efficient modules, while maintaining reasonable generalization and global context modeling via attention. 
For the edge, we also increasingly see differentiated memory modeling with newer models, like Titans \citep{behrouz_titans_2024} and B’MOJO \citep{zancato_bmojo_2025}, segmenting memory into short-term, long-term, and permanent storage, assigning specialized mechanisms to each.

In the mid-size regime, hybrids like Jamba \citep{lieber_jamba_2024} show promise, though they remain a minority and do not outperform well-tuned transformers. Their advantages are domain-specific, tied to scenarios where efficiency provides tangible gains. In general, the maturity of transformer infrastructure also makes switching to other architectures costly due to ecosystem inertia \citep{inertia}\@. However, work that enables the conversion of pretrained transformers to alternative architectures without retraining, such as RWKV, starts lowering these barriers. 

Regarding the types of hybrids we see, striped and fusion, there is no clear tendency in current research to use one over the other, since this choice highly depends on what primitives are combined. Using full attention in a fusion hybrid comes with no gains in efficiency, while combinations of purely subquadratic primitives can benefit from fusion to balance out their different disadvantages compared to full attention.

Together, these trends signal a shift toward architectural diversity. While transformers remain dominant, alternatives are finding footholds in specific use cases and operational niches.
\subsection{Outlook}
\label{subsection:Outlook}
At the frontier, full attention is likely to remain central for the foreseeable future. Still, even these models may begin incorporating hybrid elements, especially for memory management or task-specific routing. In this sense, we also anticipate model routing and \emph{Mixture of Architectures} (MoA) paradigms to become more relevant. The shift is not toward replacement, but toward building flexible systems from a growing set of specialized primitives, an idea that has already been surfaced by \citet{yu_model_routing}, \citet{varangot_model_routing} and \citet{fu2024moa}, and continues to gain traction.

Several new open source models that interleave sparse and global attention (so $\mathcal{O}(n^2)$ hybrids) were released between the initial and camera-ready submission of our work. As of 2026-05-13, they are in the very top of both the \textit{WebDev} (up to top 5) and \textit{Text} (up to top 20) Chatbot Arena leaderboards, reinforcing our expectations: Some examples are Mimo-v2.5-pro \citep{mimo2026v25pro} and Gemma-4-31b \citep{google_gemma4_2026} that both use sliding window attention combined with global attention, GLM-5.1 \citep{GLM5}, which switches from global to sparse attention after mid-training, and Deepseek-V4 \citep{deepseekai2026deepseekv4}, which uses a striped hybrid architecture comprising compressed forms of sparse and global attention.
\section{Conclusion}
\label{section:Conclusion}
Through our review of recent subquadratic architectures, we have highlighted the most promising alternatives to full attention for sequence modeling in NLP\@. Our analysis shows that sub-quadratic elements introduce valuable tradeoffs in efficiency and latency, particularly in edge and mid-sized deployments, but do remain fundamentally constrained in generality compared to transformers. We do not expect pure subquadratic architectures on the frontier for the foreseeable future, but see Hybrids catching up fast. 

\section*{Limitations}
As a focused and concise survey, our work comes with several limitations. We restrict our analysis to language models, and therefore, our findings may not generalize to other modalities such as vision, audio, or multimodal systems. Additionally, the performance comparison presented in Table~\ref{tab:benchmarks} is limited in its language coverage, as it focuses primarily on English. There is also a slight variation in training data and procedure across the benchmark results of the models we report on, which is explained in Section~\ref{sec:benchmarking_details}. Finally, while aimed at identifying and synthesizing all relevant literature, researchers with a different focus could consider some missing works more significant.

\section*{Acknowledgments}

All analysis, research, and ideas are either our own or cited. This work used LLM-based tools for language edits and clarity improvements. This research has been funded by the German Federal Ministry of Research, Technology, and Space (BMFTR) through grant 01IS23069 Software Campus 3.0 (Technical University of Munich) as part of the Software Campus project “Know ELViS”.

\bibliography{acl_latex}
\bibliographystyle{acl_natbib}

\clearpage 
\appendix
\section{Appendix}
\label{sec:appendix}

\subsection{Survey Methodology}
\label{sec:methodology}

Our survey followed a two-fold methodology: First, to determine which alternative model architectures to include, we began with a set of seed papers drawn from recent articles in the field, namely \citet{wang_beyond_2024}, \citet{mamba_gu_23}, \citet{sun2023retentive}, and \citet{tay_efficient_2022}. From this base, we employed a backward and forward snowballing strategy: we examined the references cited within these seed papers (backward snowballing) as well as subsequent papers that cited them (forward snowballing). This iterative process enabled us to trace the development and recurrence of specific architectural primitives over time and across various research communities. Architectures that consistently reappeared in recent high-impact publications were included in the main body of our review. 
Architectures with limited recurrence and marginal impact were excluded.

Second, for the chapter discussing the fundamental limitations of quadratic and sub-quadratic architectures, we conducted a systematic literature review. This involved querying several academic databases with the search term

\emph{("fundamental limitation") AND ("transformer" OR "attention" OR "subquadratic") AND ("natural language processing" OR "NLP" OR "language model")}

to identify relevant theoretical and empirical work. The results, i.e., number of hits for each platform, and the search space (full text or abstract only), are stated in the following:

\begin{itemize}
    \item ACL: 300 (full text)
    \item Semantic Scholar: 258 (full text)
    \item Google Scholar: 4430 (full text)*
    \item IEEE: 4 (abstract)
\end{itemize}

We then condensed our findings and reported on the very core of limitations that the other findings build upon. Secondary limitations were moved to Appendix~\ref{sec:additional_limitations}. *For Google Scholar, we used additional filtering to address the high number of hits and relatively low overall relevance. The SLR cutoff was 2025-06-18, but we continued to include relevant individual papers until the paper submission.

\subsection{Honorable Mentions}
\label{sec:honorable_mentions}
In our work, we have encountered various interesting and previously impactful subquadratic architectures, which, however, we were not able to include in the main body of this paper. This was usually due to a combination of limited space and our findings that these architectures were outperformed by others before they became relevant in the long run. For completeness, this section gives a brief overview of these works.

\begin{itemize}
\setlength\itemsep{0em}

    \item \textbf{Extended Long Short-Term Memory (xLSTM)}
    xLSTM \citep{beck_xlstm_2024} enhances the LSTM architecture by incorporating state expansion, normalization, and stabilization techniques and also using exponential gating and matrix memory. It stacks two specialized LSTM modules: sLSTM, with scalar memory and update mechanisms for efficient state mixing and tracking, and mLSTM, with matrix memory and a covariance-based update rule for improved memorization and parallelism. The mLSTM’s matrix memory supports tasks like Multi-Query Associative Recall. 
    xLSTM has linear time and constant memory complexity, but incurs overhead from complex memory operations, only partially offset by hardware-aware optimizations.
    \item \textbf{HGRN}
    The Hierarchically Gated Recurrent Neural Network (HGRN) \citep{qin_hierarchically_2023} consists of stacked layers comprising token-mixing (HGRU) and channel-mixing (GLU) modules. Unlike S4 or RWKV-4, HGRN also uses data-dependent, dynamic decay rates via forget gates, allowing lower layers to focus on short-term and higher layers on long-term dependencies. Learnable lower bounds on forget gates prevent vanishing gradients. To address limited recurrent state size, HGRN2 \citep{qin_hgrn2_2024} expands the state non-parametrically, improving scaling and outperforming Mamba on Long Range Arena \citep{tay_long_2021}, though pretrained transformers like LLaMA \citep{touvron_llama_2023} still perform better on long-context tasks. HGRN2 has been scaled to 3B parameters.
\end{itemize}

\subsection{Benchmark Score Sourcing}
\label{sec:benchmarking_details}
The exact methodology through which benchmark scores are reported on in the literature referenced throughout this survey varies strongly. Table \ref{tab:benchmarks} represents our best effort at consolidating these results without introducing yet another evaluation suite. We chose to use the scores from \citet{qwen2025qwen25technicalreport} for Qwen2.5 and Llama 3.1, and \citet{peng_rwkv-7_2025} for Llama 3.2 and RWKV7, due to their consistent evaluation approach. For all other models, we gathered the results from their original technical papers, ensuring consistency to the best of our knowledge. Nevertheless, some inconsistencies, namely in the number and type of tokens used during training, and differences in the number of shots for some task/model combinations, remain.

The model references are as follows: Titans \citep{behrouz_titans_2024}, Griffin \citep{de_griffin_2024}, GLA \citep{yang_gatedLinearAttention_2024}, HGRN2 \citep{qin_hgrn2_2024}, Mamba2 \citep{mamba2_dao_24}, xLSTM \citep{beck_xlstm_2024}, BMoJo \citep{zancato_bmojo_2025}, RWKV7 \citep{peng_rwkv-7_2025}, Samba \citep{ren_samba_2025}, Jamba \citep{lieber_jamba_2024}, Qwen2.5 \citep{qwen2025qwen25technicalreport}, Llama3.1 \citep{grattafiori2024llama}, Mixtral \citep{jiang2024mixtral}. 

The references for the benchmarks are MMLU \citep{MMLU}, Lambada \citep{paperno2016lambada}, PIQA \citep{bisk2020piqa}, BBH \citep{bbh}, ARC-E and ARC-C \citep{arc-c}, Winogrande \citep{winogrande}, HellaSwag \citep{zellers-etal-2019-hellaswag}, GSM8k \citep{gsm8k}, and HumanEval \citep{humaneval}, all used scores are accuracy based.

\subsection{Additional Limitations of Attention}
\label{sec:additional_limitations}
In this section, we list interesting additional limitations that were not included in the main body of the paper.

\begin{itemize}
\item \citet{hahn-2020-theoretical} prove that pure attention Transformers cannot handle bracket matching, iterated negation, or non-counter-free regular languages on long inputs, nor emulate stacks or arbitrary finite-state automata (unless layers or heads scale with input length).

\item \citet{sanford_representational_2023} show that single-layer, multi-head Transformers require polynomially more heads or dimensions to solve certain triple detection tasks, and likely struggle with higher-order tasks like Match3 \citep{sanford_representational_2023} without hints or augmentation. However, most real-world sequence problems decompose into pairwise relationships, aligning well with transformer capabilities. 

\item \citet{huang2025_generalization} propose a theoretical framework to investigate length generalization in causal transformers that use learnable absolute positional encodings. By introducing constraints on how positional information can be utilized, their framework allows them to derive results for multilayer models. They formally prove problems with poor length generalization, such as copying sequences containing repeated strings. Although it remains an open question whether the expressivity of transformers goes beyond the complexity class $TC^0$, their findings suggest a potential distinction between problems solvable within $TC^0$ and those for which length generalization is feasible with absolute positional encodings.

\item \citet{amiri_lowerboundscot_2025} investigate systematic lower bounds on the number of CoT steps required for various algorithmic problems within a hard-attention setting. Their analysis demonstrates that the required CoT length must necessarily scale with the input length, thereby constraining the ability of self-attention models to solve these tasks efficiently with small inference-time compute.

\item \citet{peng_limitationstransformer_2024} prove that a single transformer layer is not able to do function composition if the domain size of the functions is larger than the dimension parameters of the transformer. Moreover, they show that if we leverage CoT, the model needs to generate a $\Omega(\sqrt{n})$ long prompt to solve iterated function composition, with $n$ being the number of tokens in the prompt. They assume that multi-layer transformers struggle as well.

\item \citet{saparov_search_2025} argue that transformers with standard training will not have robust searching and planning abilities, no matter their number of parameters. For small graphs, a model with effectively limitless and idealized training data can learn to search. Nevertheless, according to them, even if a model can use search in-context (i.e., CoT), it still struggles with search on larger graphs.

\item \citet{han_bridging_2025} show that linear attention is not injective, often assigning identical attention weights to different queries and causing semantic confusion. They also demonstrate that linear attention struggles with effective local modeling, a strength of softmax attention. Moreover, the low-rank nature of linear attention’s feature map can further hinder modeling of complex spatial or local information \citep{fan_breaking_2025}.
\end{itemize}

\subsection{Modern \texorpdfstring{\boldmath$\mathcal{O}(n^2)$}{Quadratic} Attention}\label{subsection:moderno2}
The core principle of quadratic attention has not changed much in recent years. However, system-level improvements significantly influence the discussion and use of attention today. Although these methods are not the focus of our work since they do not change the $\mathcal{O}(n^2)$ bottleneck, many attention variants deliver substantial practical speedups with no reduction in quality compared to standard attention.  We briefly cover the most common techniques to establish a fair context for the later discussion of alternative architectures.
\paragraph{KV Cache Optimizations}
During inference, attention's keys and values are often cached to avoid redundant computation, making efficient \emph{key-value} (KV) cache management key for reducing memory requirements. \emph{Multi-Query Attention} (MQA) \citep{shazeer_fast_2019} and \emph{Grouped-Query Attention} (GQA) \citep{ainslie_gqa_2023} share key and value matrices across attention heads, reducing cache size by a constant factor at the cost of reduced expressivity. \emph{Multi-Head Latent Attention} (MLA), introduced by DeepSeek \citep{deepseek-ai_deepseek-v2_2024, deepseek-ai_deepseek-r1_2025}, shares a latent matrix among heads, which is projected back individually, achieving similar cache savings with better performance than MQA and GQA\@. Refer to \citet{li_survey_2024} and \citet{shi_keep_2024} for a detailed overview of KV cache techniques.

The \emph{Paged Attention} algorithm \citep{kwon_efficient_2023} enables the storing of attention keys and values in non-contiguous paged memory. More specifically, it improves inference memory efficiency by partitioning the KV cache into fixed-size pages and tracking them via a page table, boosting throughput 2–4× and eliminating padding.
\paragraph{Flash Attention}
FlashAttention \citep{flash_attention_dao_22} and its successors 
exploit GPU memory hierarchies to make attention both faster and more memory-efficient, reducing memory usage to be linear in sequence length and delivering 2–4× runtime speedups over strong baselines. FlashAttention-2 \citep{flash_attention2_dao_23} improved thread work partitioning for further speedup (as proven by GPT-style \citep{gpt3} LLM training), while FlashAttention-3 \citep{shah_flashattention-3_2024}, specialized for Hopper GPUs, adds asynchrony and low-precision operations for an additional 1.5–2× boost.
\paragraph{Forgetting Attention}
The \textit{Forgetting Transformer} (FoX) \citep{lin2025forgetting} modifies standard softmax attention by adding a learned forget gate that controls how strongly past tokens remain available. Instead of treating all previous context equally persistently, it applies a decay factor so that older or less relevant information gradually fades. FoX can improve long
context language modeling and length extrapolation, but still computes the full attention matrix, so it remains quadratically scaling with context length. Forgetting Attention is compatible with the FlashAttention algorithm.

\end{document}